\documentclass[10pt,twocolumn,letterpaper]{article}

\usepackage{iccv}
\usepackage{times}
\usepackage{epsfig}
\usepackage{graphicx}
\usepackage{amsmath}
\usepackage{amssymb}
\usepackage{color}
\usepackage{bm}
\graphicspath{{./figures/}}

\usepackage[pagebackref=true,breaklinks=true,letterpaper=true,colorlinks,bookmarks=false]{hyperref}

\iccvfinalcopy 

\def\iccvPaperID{****} 

\ificcvfinal\fi
\begin{document}

\title{DelugeNets: Deep Networks with Efficient and Flexible Cross-layer Information Inflows}

\author{Jason Kuen$^1$\\\
	{\tt\small jkuen001@ntu.edu.sg}
	\and
	Xiangfei Kong$^1$\\
	{\tt\small xfkong@ntu.edu.sg}
	\and
	Gang Wang$^2$\\
	{\tt\small gangwang6@gmail.com}
	\and
	Yap-Peng Tan$^1$\\
	{\tt\small eyptan@ntu.edu.sg}
	\and
	Nanyang Technological University$^1$\quad Alibaba Group$^2$
}

\maketitle

\begin{abstract}
	Deluge Networks (DelugeNets) are deep neural networks which efficiently facilitate massive cross-layer information inflows from preceding layers to succeeding layers. The connections between layers in DelugeNets are established through cross-layer depthwise convolutional layers with learnable filters, acting as a flexible yet efficient selection mechanism. DelugeNets can propagate information across many layers with greater flexibility and utilize network parameters more effectively compared to ResNets, whilst being more efficient than DenseNets. Remarkably, a DelugeNet model with just model complexity of 4.31 GigaFLOPs and 20.2M network parameters, achieve classification errors of 3.76\% and 19.02\% on CIFAR-10 and CIFAR-100 dataset respectively. Moreover, DelugeNet-122 performs competitively to ResNet-200 on ImageNet dataset, despite costing merely half of the computations needed by the latter.
	\vspace{-0.1cm}
\end{abstract}

\section{Introduction}\label{introduction}
Deep learning methods \cite{bengio2009learning,schmidhuber2015deep}, particularly convolutional neural networks (CNN) \cite{lecun1998gradient} have revolutionized the field of computer vision. CNNs are integral components of many recent computer vision techniques which spread across a diverse range of vision application areas \cite{gu2015recent}. Hence, developing more sophisticated CNNs has been a prime research focus. Over the years, many variants of CNN architectures have been proposed. Some works focus on improving the activation functions \cite{he2015delving,xu2015empirical}, and some focus on increasing the heterogeneity of convolutional filters within the same layers \cite{szegedy2015going,szegedy2016rethinking}. Lately, the idea of improving CNNs by greatly deepening them has gained much traction, following the immense successes of Residual Networks (ResNets) \cite{he2016deep,he2016identity} in image classification.

ResNets make use of residual connections to support relatively unobstructed information flows (shortcut) between layers. Each succeeding layer receives the sum of all its preceding layers’\footnote{The unit layer in ResNet, ResNet-like, and DenseNet models refers to a \textit{composite layer}  formed by several basic layers. See Section \ref{CL}.} outputs as input. Compared to traditional non-residual deep networks, outputs of preceding layers in ResNets can reach succeeding layers with minimal obstructions, even if the preceding layer and succeeding layer is separated by a very long layer-distance. However, the cross-layer connections between preceding and succeeding layers of ResNet are fixed and not ``“selective", and therefore the succeeding layers are not able to prioritize or deprioritize output channels of certain preceding layers. Instead, the outputs of preceding layers are lumped together via simplistic additive operation, making it very tough for succeeding layers to perform layer-wise information selection. The inflexibility of residual connections also hinders the ability of ResNets to learn cross-layer interactions and correlations.

\begin{figure*}[ht]
	\begin{center}
		\includegraphics[width=0.91\textwidth]{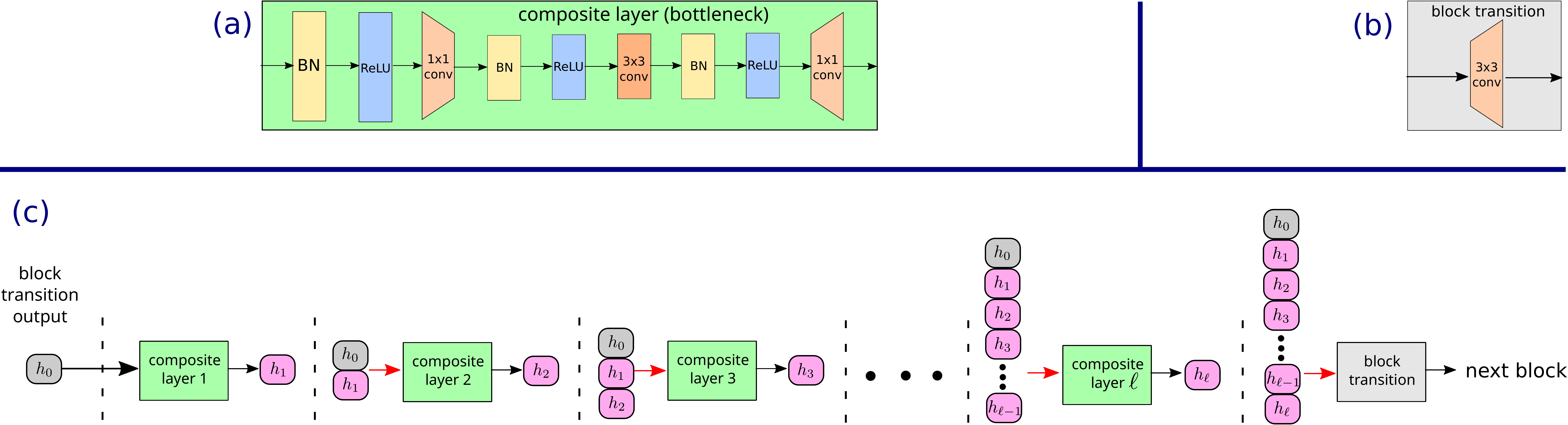}
	\end{center}
	\caption{Deluge Network components: (a) a composite layer, (b) a block transition component, and (c) a block. \textcolor{red}{Red}-colored arrows indicate $1\times1$ cross-layer depthwise convolutions.}
	\label{fig:delugenets}
	\vspace*{-0.1cm}
\end{figure*}

Densely connected networks (DenseNets) \cite{huang2016densely} aim to overcome this drawback of ResNets, by having convolutional layers to consider an extra dimension - the depth/cross-layer dimension, in addition to the spatial and feature channel dimensions used in regular convolutions. In DenseNets, the input feature maps to succeeding layers are concatenations of preceding layers’ outputs, rather than simple summations. Hence, when applying convolution operations on the concatenated feature maps, 
the convolutional filters have to learn spatial, cross-channel, and cross-layer correlations altogether, entailing heavy amounts of parameters (filter width $\times$ filter height $\times$ \# input channels $\times$ \# output channels $\times$ \# preceding layers) and computations. DenseNet-BC \cite{huang2016densely} was recently introduced as a more efficient variant of DenseNet, in which the filters have to consider just cross-channel and cross-layer correlations. Despite that, considering that DenseNets' composite layers receive inputs from several dozens of preceding layers, the computation and parameter requirements are still rather high.

To counter excessive computational complexity and parameter growth, DenseNet models are specifically configured to have much lower output \textit{width} (between 12 and 48 output channels) at each layer, compared to typical image classification CNNs. However, it is crucial to have considerable network width as contended by \cite{zagoruyko2016wide}, and decreasing output width too much is harmful to network’s representational power. Furthermore, by visualizing DenseNet's weight norms, Huang et al. \cite{huang2016densely} showed that the features of preceding \textit{composite layers} get reused directly by the succeeding \textit{composite layers} in a rather infrequent manner. Yet, these ``diminished" features have to be processed by relatively expensive convolution operations in DenseNets.

Thus, in this paper, we propose a new class of CNNs called Deluge Networks (DelugeNets) which enable \textbf{flexible cross-layer connections} yet have \textbf{regular output width} in each composite layer. As a result of using regular output width, the information inflows from preceding layers to succeeding layers in DelugeNets are massive, in contrast to the lesser information inflows in DenseNets.  DelugeNets are inspired by separable convolutions \cite{jin2014flattened,jaderberg14speeding,chollet2016deep,lin2014network}. The efficiency of convolutions can be improved by separating the combined dimensions involved, resulting in separable convolutions. DelugeNets are designed such that the depth/cross-layer dimension is processed independently from the rest (channel and spatial dimensions), using a novel variant of convolutional layer called \textit{cross-layer depthwise convolutional layer} (see Figure \ref{fig:CDC}) as described in Section \ref{CDC}. Cross-layer depthwise convolutional layers handle only \textbf{cross-layer} interactions and correlations, without getting burdened by other dimensions. They facilitate cross-layer connections in DelugeNets in a very efficient and effective manner. Experiments show the superior performances of DelugeNets in terms of classification accuracy, parameter efficiency, and more remarkably computational complexity.

\section{Related Work}\label{relatedwork}
\subsection{Training Deep Networks}
Developing methods for training very deep neural networks is a rather significant research topic that has received much attention over the years. Lee et al. \cite{lee2015deeply} incorporate classification losses into intermediate hidden layers, allowing unimpeded supervised signals to reach the layers. In a similar spirit as \cite{lee2015deeply}, GoogleNets \cite{szegedy2015going} and Inception \cite{szegedy2016rethinking} models attach auxiliary classifiers to a few intermediate layers to encourage feature discriminativeness in lower network layers. DelugeNets, by contrast, can readily backpropagate the supervised signals to earlier layers without relying on additional losses, due to connections supporting flexible information inflows from preceding to succeeding layers.

There is another stream of works focusing on improving the information flows between layers of very deep networks, which is also the focus of our work. Highway Networks \cite{srivastava2015training, greff17highway} make use of a Long-Short-Term-Memory (LSTM \cite{hochreiter1997long})-inspired gating mechanism to control information flow from linear and nonlinear pathways. Through appropriately learned gating functions, information can flow unimpededly across many layers, which can be thought of as a kind of flexible mechanism to combine cross-layer information inflows. He et al. \cite{he2016deep} propose Residual Networks (ResNets) which compute the residual (additive) functions of the outputs of linear and nonlinear pathways, without complex gating mechanisms. ResNets have shown to tackle well the vanishing gradient and network degradation problems that occur in very deep networks. The pre-activation variants of ResNet (ResNet-v2) \cite{he2016identity} normalize incoming activations at the beginnings of residual blocks to improve information flow and regularization. 

Instead of going deeper, Wide-ResNets \cite{zagoruyko2016wide} improve upon originally proposed ResNets by having more convolutional filters/channels (width) and less numbers of layers (depth). Motivated by the high model complexity of ResNets in terms of depths and parameter numbers, several ``dropping"-based regularization methods \cite{huang2016deep,singh2016swapout} have been developed for regularizing large ResNet models. ResNets can be seen as a less flexible special case of DelugeNets, in which the cross-layer connection weights are not learnable and fixed as ones. Densely connected networks (DenseNets) \cite{huang2016densely} which we discuss extensively in Section \ref{introduction} belong to the same stream of works.

\subsection{Separable Convolutions}
Separable convolutions have been adopted to construct efficient convolutional networks. Earlier works \cite{jaderberg14speeding,denton2014exploiting} compress convolutional networks by finding low-rank approximation of convolutional layers of pre-trained networks. Network-in-network \cite{lin2014network} employs $1\times1$ pointwise (cross-channel) convolutional layers to enrich representation learning in an efficient manner. $1\times1$ pointwise convolutions are generally coupled with other convolution variants (e.g., spatial convolutions) to achieve separable convolutions. Flattened convolutional networks \cite{jin2014flattened} are equipped with one-dimensional convolutional filters of 3 dimensions (channel, horizontal, and vertical) which are processed sequentially and trained from scratch. For maximal channel-spatial separability, a conventional convolutional layer can be replaced with depthwise separable convolution (spatial depthwise convolution followed by a $1\times1$ pointwise convolution), as demonstrated by Xception \cite{chollet2016deep}. In contrast to these existing works which mainly deal with channel-spatial separability, the work in this paper deals with ``cross-layer"-channel separability. Also, to the best of our knowledge, this paper is the first work on cross-layer depth/channelwise convolutions.

\section{Deluge Networks}
Similar to existing CNNs (ResNet \cite{he2016deep,he2016identity}, VGGNet \cite{simonyan2015very}, and AlexNet \cite{krizhevsky2012imagenet}), DelugeNets gradually decrease spatial sizes and increase feature channels of feature maps from bottom to top layers, with a linear classification layer attached to the end. The layers operating on the same feature map dimensions can be grouped to form a \textit{block}. In DelugeNets, the input to a particular layer comes from \textbf{all of its preceding layers} of the same \textit{block}. There is no information directly flowing from other \textit{blocks}. Within any \textit{block}, the cross-layer information flows through connections established by the cross-layer depthwise convolutions (see Section \ref{CDC}). For transition to the next \textit{block} as described in Section \ref{transition}, we perform cross-layer depthwise convolution followed by strided spatial convolution to obtain feature map of matching dimensions. The structure of \textit{block} in DelugeNets is illustrated in Figure \ref{fig:delugenets}(c), with individual layers separated by vertical dashed lines.

\subsection{Composite Layer}\label{CL}
In CNNs, a \textit{layer} often refers to a \textit{composite layer} of several basic layers such as Rectified Linear Unit (ReLU), Convolutional (Conv), Batch Normalization (BN) layers. Inspired by \cite{he2016identity}, we use the bottleneck-kind of \textit{composite layer} \underline{BN-ReLU-Conv-BN-ReLU-Conv-BN-ReLU-Conv} in DelugeNets, as illustrated in Figure \ref{fig:delugenets}(a). This kind of \textit{composite layer} is designed to improve parameter efficiency in deep networks, by employing $1\times1$ spatial convolutional layers at the beginning to reduce channel dimension, and at the end to increase channel dimension. In the ResNet models proposed by \cite{he2016identity}, the base channel dimensions are increased by 4 times. We however only increase them by 2 times in this paper, for the reason that we can allocate more computational and parameter budgets to train deeper DelugeNets.

Such a \textit{composite layer} has also shown to work very well for very deep neural networks which combine information from multiple sources, such as ResNets and the proposed DelugeNets. The primary reason that it works well is that combined multi-source information is normalized via BN layers before it is passed into the upcoming weight (convolutional) layers. This reduces internal covariate shift and regularizes the model more effectively \cite{he2016identity} than just passing unnormalized multi-source information to the weight layers.

\begin{figure}[ht]
	\centering
	\includegraphics[width=0.8\linewidth]{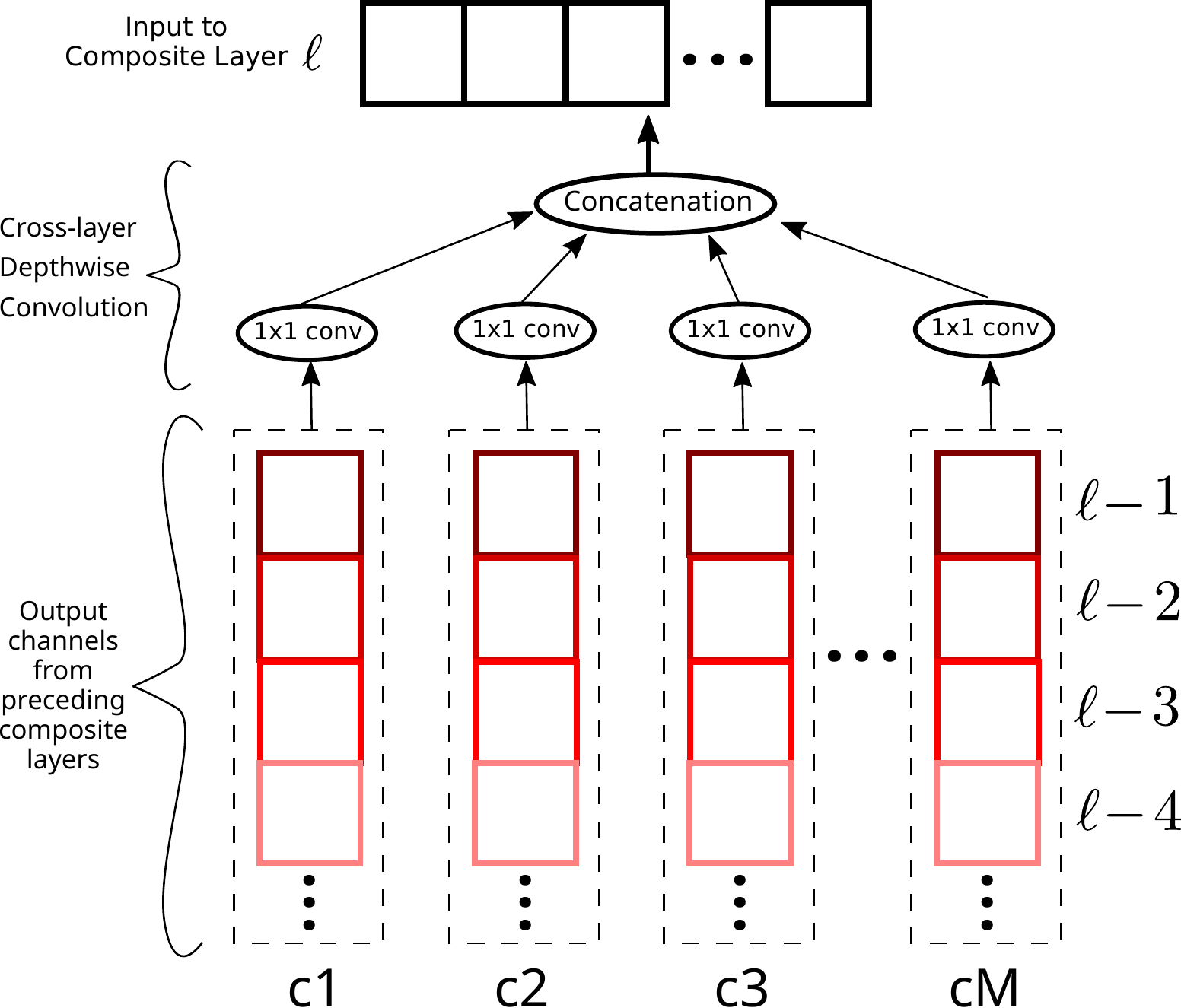}
	\caption{Cross-layer Depthwise Convolution. The columns correspond to feature channel indices, and the rows correspond to preceding \textit{composite layer} indices.}
	\label{fig:CDC}
	\vspace*{-0.1cm}
\end{figure}

\subsection{Cross-layer Depthwise Convolutional Layers}\label{CDC}
To facilitate efficient and flexible cross-layer information inflows, in this paper, we develop a cross-layer depthwise convolution method. A cross-layer depthwise convolutional layer concatenates the channels of feature map outputs of many layers, and then applies (channel,spatial)-independent filters to the concatenated channels. Equipped with such filters, DelugeNets are able to process the depth/layer dimension independently of the rest (channel and spatial dimensions), as mentioned in Section \ref{introduction}. Figure \ref{fig:CDC} gives a graphical illustration of cross-layer depthwise convolution operation.

Cross-layer depthwise convolutional layers facilitate the inflows of information from preceding \textit{\textit{composite layer}s} to succeeding \textit{\textit{composite layer}s}. Suppose that $\ell$ denotes the layer of an arbitrary \textit{composite layer}, and $ h^c_{\ell-i}\in\mathbb{R} $ denotes the $c$-th channel of the preceding $(\ell-i)$-th \textit{composite layer}'s output. And, there are $ N $ number of preceding \textit{\textit{composite layer}s}, as well as one preceding block transition output $\bm{h}_0$. The $c$-th channel of the input $\bm{x}_\ell$ to \textit{composite layer} $\ell$ is:

\begin{equation}
x^c_\ell =  \sum\limits_{i=1}^{N+1} w^c_{\ell-i}h^c_{\ell-i} + b^c_\ell
\end{equation}

\noindent
where $ w^c_{\ell-i}\in\mathbb{R} $ and $ b^c_\ell\in\mathbb{R} $ are the filter weights and bias respectively, for each channel. We streamline the equations by not having spatial location-related notations, and the weights and biases are assumed to be shared across all $1\times1$ spatial locations (spatially independent) in the input feature maps as mentioned earlier.

The parameter cost of adding cross-layer depthwise convolutional layer to any existing network architecture is relatively low compared to other network parameters. For an arbitrary \textit{composite layer} in the network, the number of additional parameters is merely $ N \times M + 1 $, where $ M $ is the number of feature channels. Experimentally, we find that these extra parameters on average make up about $3\%$ of entire model parameters. In terms of computational complexity (measured in floating-point operation numbers or FLOPs), cross-layer depthwise convolutions on average cost $3\%$ more, compared to baseline models without such convolutions. DenseNets \cite{huang2016densely}, on the other hand, require heavy amounts of computations and parameters to connect to preceding layers, through cross-layer output concatenations followed by $3\times3$ or $1\times1$ spatial convolutions.


\textbf{Advantages:} Cross-layer depthwise convolutional layers are beneficial because they encourage features generated by a preceding \textit{\textit{composite layer}s} to be taken as inputs for many times by the succeeding layers (feature reuse). This naturally leads to parameter efficiency because there is no need to redundantly learn filters which generate the same features in succeeding layers, in case those features are needed again later. Furthermore, in conventional ReLU-based convolutional networks, features that get turned off by ReLU activation functions (at the beginnings of blocks) cannot be recovered by other network parts or layers. In DelugeNets, via the use of cross-layer depthwise convolutional layers, output of a preceding \textit{composite layer} can be transformed differently for each succeeding \textit{composite layer} to serve as input. Consequently, input features that get turned off at the beginning of certain succeeding \textit{\textit{composite layer}s} may be active in others.

In CNNs, the filter weights are shared across many spatial locations in the feature maps. The weight sharing mechanism acts as an effective regularizer. Similar to the CNN's weight sharing mechanism, the same features in DelugeNets' preceding \textit{composite layers} are shared by the succeeding \textit{composite layers}. As a result, the weights of \text{composite layers} in DelugeNets become more regularized. Based on such consideration, we allocate more model parameters to the spatially smaller network blocks, by setting the number of \textit{composite layer} in succeeding block to be larger than preceding block. The motivation behind this is to achieve \textbf{lower computational complexity} (since smaller feature maps are computationally cheaper to process), relying more on cross-layer feature reuse and less on parameter-sharing across spatial locations, for regularization. Such an allocation scheme differs from ResNets in which many layers and parameters are allocated for blocks with ge feature maps to regularize filters better, and less for blocks with spatially small feature maps to reduce overfitting (see Section \ref{imagenet}).

Besides encouraging feature reuse, cross-layer depthwise convolutional layers are advantageous from the perspective of gradient flow. The gradient flows in DelugeNets are regulated by multiplicative interactions with the filter weights in cross-layer depthwise convolutional layers, such that the \textit{\textit{composite layer}s} all receive unique backpropagated gradient signals even if they come from the same block. This is not true for ResNet models, in which the \textit{\textit{composite layer}s} within the same block receive identical backpropagated gradient signals, due to simple addition (residual) operation.

\begin{table*}[t]
	\begin{center}
		\resizebox{400px}{!}{%
			\begin{tabular}{l|r|r|r|r|r}
				\hline
				\textbf{Model}         & \textbf{\#Params} & \textbf{Depth} & \textbf{GigaFLOPs} & \textbf{CIFAR-10} & \textbf{CIFAR-100} \\ \hline
				Highway Network \cite{srivastava2015training}        & -        & -     & - & 7.60     & 32.24     \\ \hline
				FractalNet \cite{larsson2016fractalnet} 	           & 38.6M    & 20    & - & 4.59     & 22.85     \\ \hline
				ResNet \cite{he2016deep}                 & 1.7M     & 164   & - & 5.93     & 25.16     \\
				ResNet \cite{he2016deep}                & 10.2M    & 1001  & - & 7.61     & 27.82     \\ \hline
				ResNet with ELU \cite{shah2016deep}       & -        & 110   & - & 5.62     & 26.55     \\ \hline
				ResNet with Identity Mappings \cite{he2016identity} & 1.7M     & 164   & - & 5.46     & 24.33     \\
				ResNet with Identity Mappings \cite{he2016identity} & 10.2M    & 1001  & - & 4.62     & 22.71     \\ \hline
				ResNet with Swapout \cite{singh2016swapout} & 7.4M     & 32    & - & 4.76     & 22.72     \\ \hline
				ResNet with Stochastic Depth \cite{huang2016deep} & 1.7M    & 32    & - & 5.23     & 24.98     \\
				ResNet with Stochastic Depth \cite{huang2016deep} & 10.2M   & 1202  & - & 4.91     & -         \\ \hline
				Wide-ResNet ($04\times$width) \cite{zagoruyko2016wide}            & 8.7M     & 40    & 2.60 & 4.53     & 21.18     \\
				Wide-ResNet ($08\times$width) \cite{zagoruyko2016wide}           & 11.0M    & 16    & 3.10 & 4.27     & 20.43     \\
				Wide-ResNet ($10\times$width) \cite{zagoruyko2016wide}           & 36.5M    & 28    & 10.49 & 4.00     & 19.25     \\ \hline
				DenseNet ($k = 12$) \cite{huang2016densely}      & 7.0M     & 100   & 3.65 & 4.10    & 20.20     \\
				DenseNet ($k = 24$) \cite{huang2016densely}     & 27.2M    & 100   & 14.56 &
				3.74     & 19.25     \\
				DenseNet-BC ($k = 24$) \cite{huang2016densely}     & 15.3M    & 250   & 10.09 & 3.62     & 17.60     \\
				DenseNet-BC ($k = 40$) \cite{huang2016densely}      & 25.6M     & 190   & 18.67 & \textbf{3.46}     & \textbf{17.18}     \\ \hline
				\textbf{DelugeNet-146}          & 6.7M     & 146  & 1.43 & 3.98     & 19.72     \\
				\textbf{DelugeNet-218}          & 10.0M    & 218  & 2.13 & 3.88     & 19.31     \\
				\textbf{Wide-DelugeNet-146}     & 20.2M    & 146  & 4.31 & 3.76  & 19.02    \\ \hline
			\end{tabular}
		}
	\end{center}
	\caption{CIFAR-10 and CIFAR-100 test errors (percentage) of existing models and DelugeNets.
	}
	\label{table:cifars}
	\vspace*{-0.1cm}
\end{table*}

\subsection{Block Transition}\label{transition}
Different network blocks operate on feature maps of different spatial and channel dimensions. For block transition, there is a need to transform the feature map to match the spatial and channel dimensions of next block. In ResNet-like models, block transition can be done with either $ 1\times1 $ strided convolution, or strided average pooling with channel padding. These block transition designs aim to preserve the information from previous block by having only minimal transformation as well as dismissing any non-linear activation function. Such block transition designs are suboptimal for DelugeNets because they allow direct information flow from just the last \textit{composite layer} of the previous block, and they conceivably hinder the information flows from other \textit{\textit{composite layer}s}.

To this end, we propose a new block transition component which has a cross-layer depthwise convolutional layer followed by $3 \times 3$ spatial convolutional layer. The cross-layer depthwise convolutional layer allows direct information inflow from all \textit{\textit{composite layer}s} from the previous block, therefore summarizing the outputs of all \textit{composite layers} of previous block. Then, the $3 \times 3$ strided spatial convolutional layer (see Figure \ref{fig:delugenets}(b)) transforms the summarized feature map to have matching spatial and channel dimensions. $3 \times 3$ strided convolutional layer is chosen over $1 \times 1$ strided convolutional layer as the latter wastes the features it receives, for many of the feature map's spatial locations, while the former does not. Similar to the block transition designs in ResNets, we do not add non-linear activation functions after the spatial convolutional layer.

\section{Experiments}
To rigorously validate the effectiveness of DelugeNets, we evaluate DelugeNets on 3 image classification datasets with varied degrees of challengingness: CIFAR-10 \cite{krizhevsky2009learning}, CIFAR-100 \cite{krizhevsky2009learning}, ImageNet \cite{russakovsky2015imagenet}. The experimental code is written in Torch \cite{collobert2011torch7}, and is available at \url{https://github.com/xternalz/DelugeNets}.

\subsection{CIFAR-10 and CIFAR-100}
\textbf{Datasets:} CIFAR-10 and CIFAR-100 are 2 subsets of the Tiny Images dataset \cite{torralba2008million} annotated to serve as image classification datasets. There are 50,000 training images and 10,000 testing images for each of the 2 CIFAR datasets. For pre-processing, we subtract channel-wise means from the images, and divide them by channel-wise standard deviations. During training, data augmentation is carried out moderately as in \cite{zagoruyko2016wide,huang2016densely}, with horizontal flipping and random crops taken from images padded by 4 pixels on each side. For all CIFAR-based models, the training is carried out using a single GPU.

\textbf{Implementation:} A total of 3 different DelugeNet models are implemented and evaluated on CIFAR datasets. Similar to \cite{he2016identity,zagoruyko2016wide}, all the 3 DelugeNet models have 3 \textit{blocks} - the first block works on spatially $32\times32$ feature maps, followed by $ 16\times16 $ and $ 8\times8 $ feature maps for second and third blocks respectively. They vary only in terms of numbers of \textit{composite layers} and feature channel dimensions for the 3 blocks. To minimize manual tuning of architectural hyperparameters, we design different DelugeNet models based on a simple principle that follows the parameter allocation scheme mentioned in Section \ref{CDC} - the second block has 2 times the \textbf{numbers of \textit{composite layers}} and \textbf{feature channel dimension} (width) of the first block, the third block has 2 times of the second's, and so on:

\textbf{DelugeNet-146} has base feature channel dimensions (widths) of \{32,64,128\}, and \textit{composite layer} counts of \{8,16,24\}, for its 3 blocks (in sequential ordering) respectively.

\textbf{DelugeNet-218} shares the same base widths as DelugeNet-146, but it comes with larger \textit{composite layer} counts of \{12,24,36\} which make it a much deeper model.

\textbf{Wide-DelugeNet-146} is a $1.75\times$ wider variant of DelugeNet-146, having base widths of \{56,112,224\}, while the \textit{composite layer} counts remain the same.

The proposed models (DelugeNet-146, DelugeNet-218, and Wide-DelugeNet-146) for the 2 CIFAR datasets differ only in the numbers of output labels (10 and 100). To train the models, we run Stochastic Gradient Descent (SGD) over a total of 300 training epochs, with Nestorov Momentum \cite{sutskever2013importance} and weight decay rate of $1\mathrm{e}{-4}$. As most of the existing models we compare with in this paper do not use any dropout-like regularization, we do not use any either, for fairer comparison. The starting learning rate is 0.1, and it is decayed by factor of 0.1 at epoch 150 and 225. We set the mini-batch size as 64. All DelugeNet model parameters are initialized using He's initialization method \cite{he2015delving}, and they are trained using the same settings. The training settings are in fact identical to the settings employed in \cite{huang2016densely} to train DenseNets.

\textbf{Results:} The top-1 classification errors achieved by the DelugeNets and existing models on both CIFAR datasets are presented in Table \ref{table:cifars}. The results of existing models are obtained directly from their respective papers. As shown in Table \ref{table:cifars}, DelugeNets can benefit from ``deepening" (DelugeNet-218) and ``widening" (Wide-DelugeNet-146).

\textbf{Parameter efficiency:} DelugeNets are able to perform well despite requiring much lower numbers of learnable parameters compared to existing models. The parameter efficiencies of Delugenets are second only to DenseNet-BCs \cite{huang2016densely} which aggressively compress and reduce feature channels to save parameters. Notably, DelugeNet-218 performs competitively to Wide-ResNet ($10\times$width), on both CIFAR-10 and CIFAR-100 datasets, with merely 10M parameters compared to 36.5M parameters in Wide-ResNet. Besides, Wide-DelugeNet-146 achieves CIFAR classification errors comparable to those of DenseNet ($k=24$), with 7M fewer parameters.

\textbf{Computational complexity:} In addition to parameter numbers, we report the model complexities of DelugeNets and several other comparable models (Wide-ResNets, DenseNets, and DenseNet-BCs), in terms of floating-point operation (FLOP in giga prefix unit, Giga/GFLOP) numbers. We find that in overall DelugeNets have significantly fewer model complexities than other models. Surprisingly, DelugeNet-218 requires just $1/5$ of the FLOPs required by Wide-ResNet ($10\times$width) to achieve similar classification errors. Although DelugeNets cannot exactly match or outperform DenseNet-BC, they (DelugeNets) can achieve appreciable classification errors which are rather close to those of DenseNet-BCs, at fractions of DenseNet-BCs' complexity costs. The lower model complexities of DelugeNets are attributed to the parameter allocation scheme (Section \ref{CDC}) as well as the capability of cross-layer depthwise convolutions at alleviating overfitting, even when spatially smaller network blocks have more parameters/layers than their spatially larger counterparts.

\begin{table}[t]
	\begin{center}
		\resizebox{\linewidth}{!}{%
			\begin{tabular}{l|r|r|r|r}
				\hline
				\textbf{Model}         & \textbf{\#Params} & \textbf{GFLOPs} & \textbf{Error} & \textbf{PDiff} \\ \hline
				ResNet-like baseline   &          &       &          \\
				- 1x1 conv shortcut    & 6.15M    & 1.33   & 21.14    & - \\
				- 3x3 conv shortcut    & 6.50M    & 1.39   & 20.84    & +0.30 \\
				\textbf{DelugeNet-146}          & 6.69M    & 1.43   & \textbf{19.72}  & +1.12 \\ \hline
				ResNet-like baseline   &          &       &          \\
				- 1x1 conv shortcut    & 9.26M    & 1.98   & 20.78 & -    \\
				- 3x3 conv shortcut    & 9.55M    & 2.05   & 20.31 & +0.47  \\
				\textbf{DelugeNet-218}          & 10.00M    & 2.13   & \textbf{19.31} & +1.00  \\ \hline
				ResNet-like baseline   &          &       &          \\
				- 1x1 conv shortcut    & 18.76M    & 4.05   & 19.79 & -    \\
				- 3x3 conv shortcut    & 19.82M    & 4.25   & 19.98 & -0.19   \\
				\textbf{Wide-DelugeNet-146}     & 20.19M    & 4.31   & \textbf{19.02} & +0.77 \\ \hline
			\end{tabular}
		}
	\end{center}
	\caption{Comparison with ResNet-like baselines on CIFAR-100 test errors. The fourth column reports the performance differences (\textbf{PDiff}) between baselines and DelugeNets.}
	\label{table:baselines}
	\vspace*{-0.1cm}
\end{table}

\textbf{Ablation study:} In this work, we propose cross-layer depthwise convolutional layer and a new kind of block transition design with $3\times3$ spatial convolution, which differentiate DelugeNets from existing networks. To better understand the contributions of these components, we construct ResNet-like baselines on the 3 proposed DelugeNet models. There are 2 types of baselines for each of the DelugeNet models: The \textbf{first baseline} has all of its cross-layer depthwise convolutions replaced by residual connections. Alternatively, the residual connections can be seen as cross-layer depthwise convolutional layers, whose weights are fixed as ones as pointed in Section \ref{relatedwork}. The \textbf{second baseline} is similar to the first one except that it is equipped with $3\times3$ convolutional shortcuts for block transitions, similar to our proposed block transition design. Other than those mentioned, all aspects of the baselines and their corresponding DelugeNets are the same, including training settings. We evaluate the baselines on CIFAR-100. The results are shown in Table \ref{table:baselines}.

Block transitions with $3\times3$ convolutional shortcuts can mildly improve the performances of DelugeNet-146's and DelugeNet-218's baselines. However, there is slight overfitting ($19.79\%\rightarrow19.98\%$) from adding $3\times3$ convolutional shortcuts to Wide-DelugeNet-146's baseline. The overfitting issue is greatly eased by having cross-layer depthwise convolutions in Wide-DelugeNet-146. As evidenced by the significant performance improvements (about 1\%) of DelugeNets over the baselines, the biggest contributor is cross-layer depthwise convolutional layer. Yet, the parameter costs incurred by adding these layers are very marginal. The smallest DelugeNet model, DelugeNet-146 (19.72\%) with just 6.69M parameters and complexity of 1.43 GFLOPs, suprisingly outperforms the biggest ResNet-like baseline (19.79\%) with 18.76M parameters and complexity of 4.05 GFLOPs. Furthermore, just tiny increases in complexity (about 3\%) are needed by cross-layer depthwise convolutions to achieve considerable performance gains. These findings reaffirm the advantages of the proposed cross-layer depthwise convolutional layer for deep networks.

\begin{figure*}[!htb]
	\def\arraystretch{0.3}
	\centering
	\begin{tabular}{ccc}
		\underline{Block 1 to Block 2} &
		\underline{Block 2 to Block 3} &
		\underline{Block 3 to Classification} \\\\
		\multicolumn{1}{l}{\textbf{CIFAR-100}} & & \\
		\includegraphics[width=.3\textwidth]{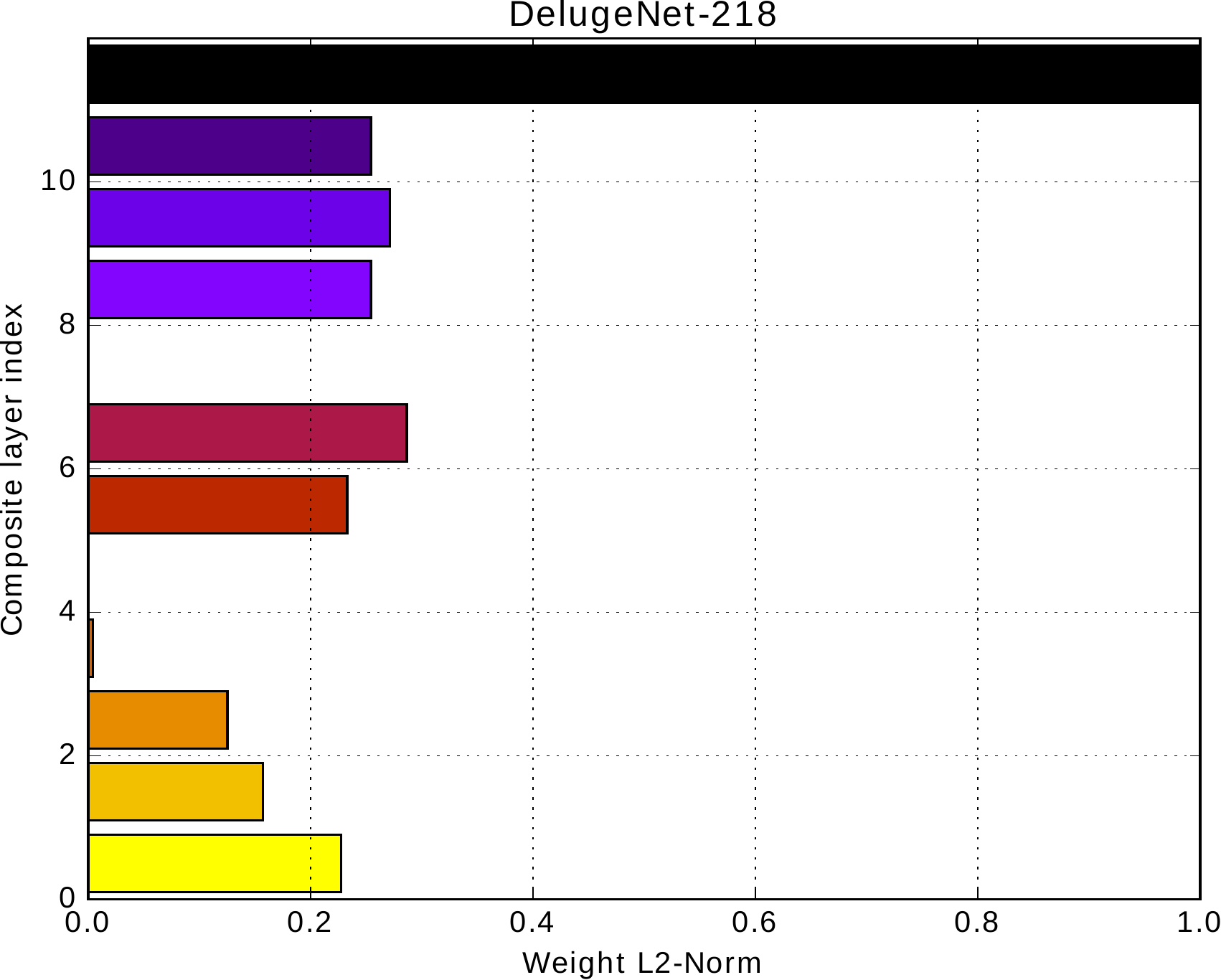} &
		\includegraphics[width=.3\textwidth]{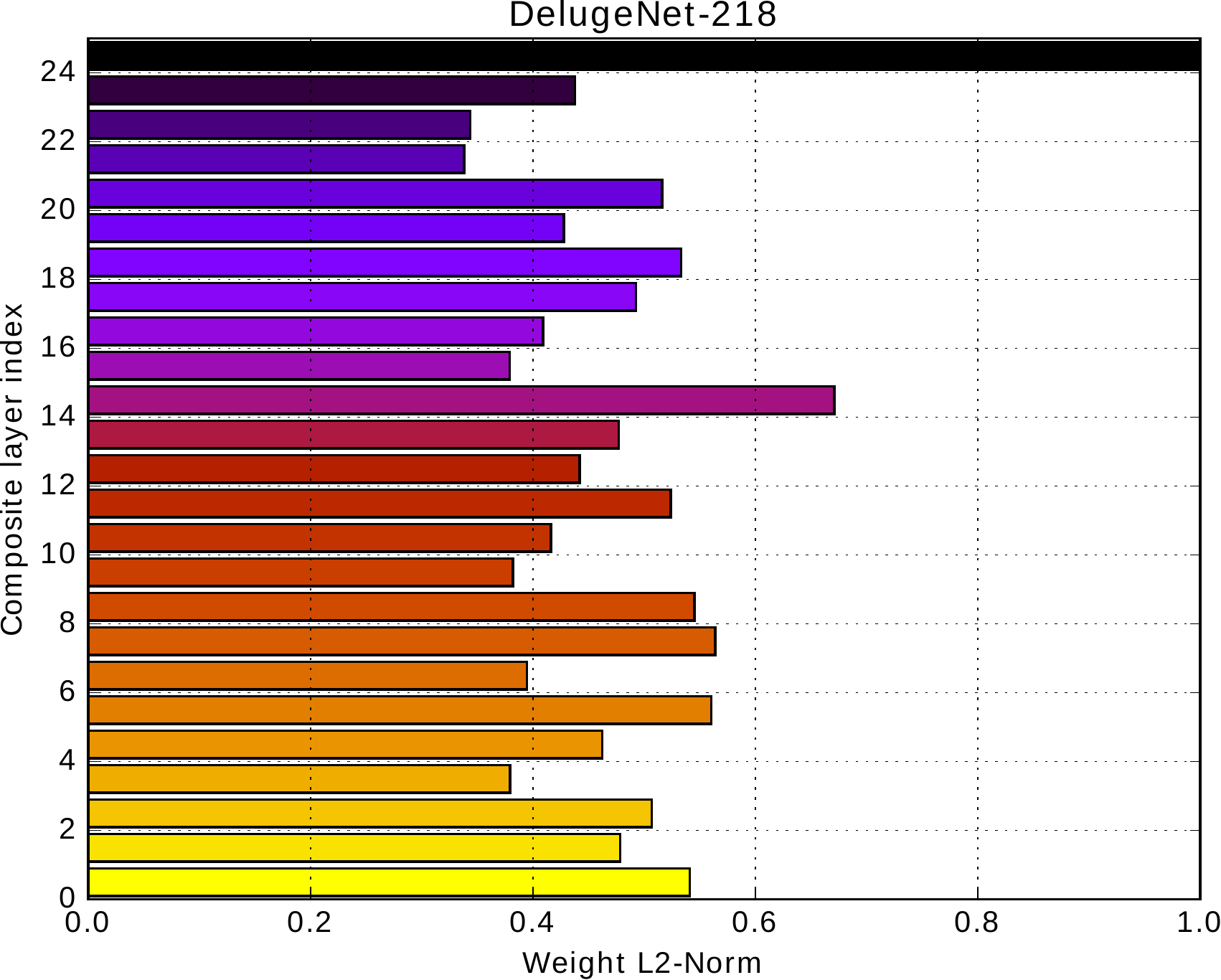} &
		\includegraphics[width=.3\textwidth]{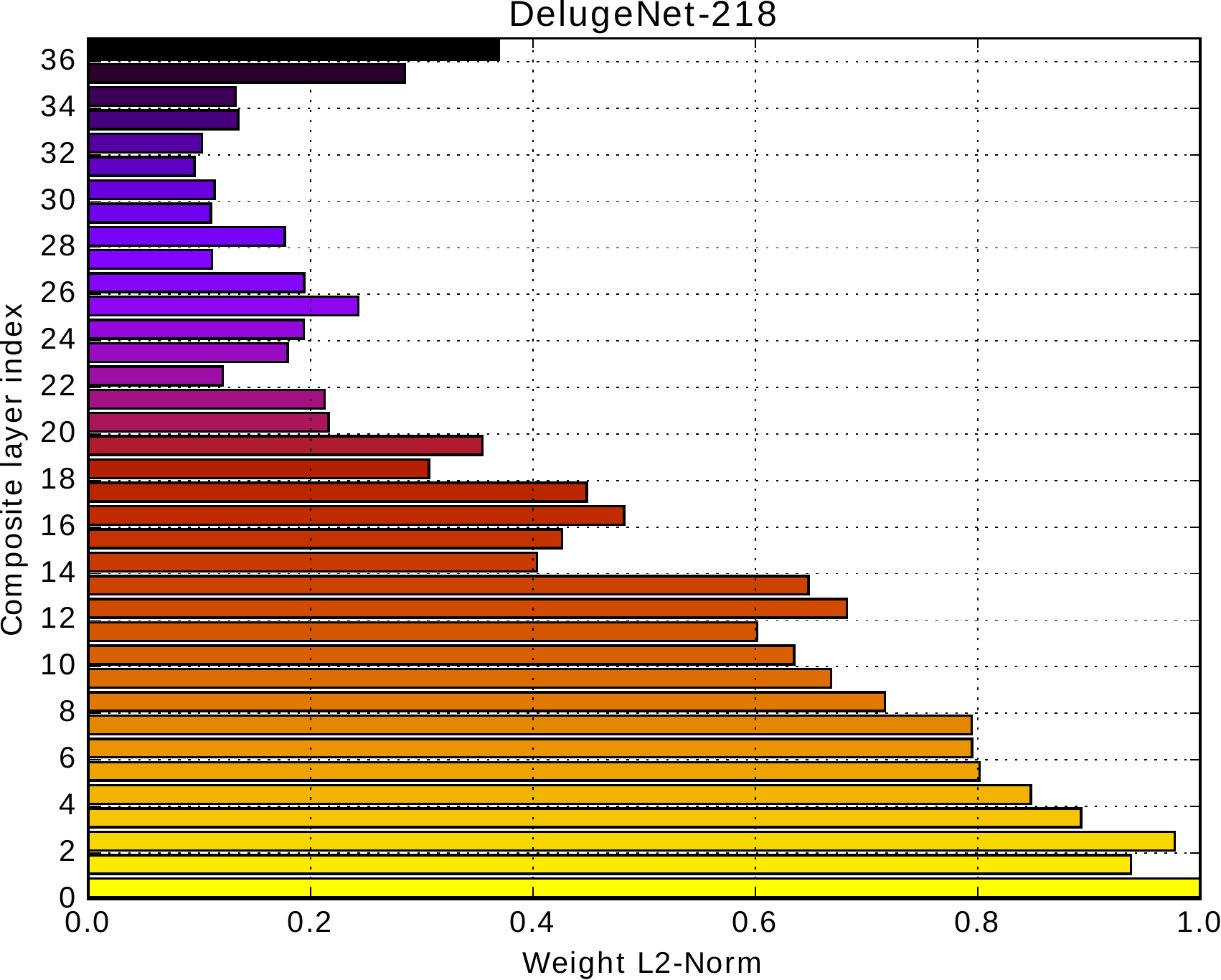} \\
		\multicolumn{1}{l}{\textbf{CIFAR-10}} & & \\
		\includegraphics[width=.3\textwidth]{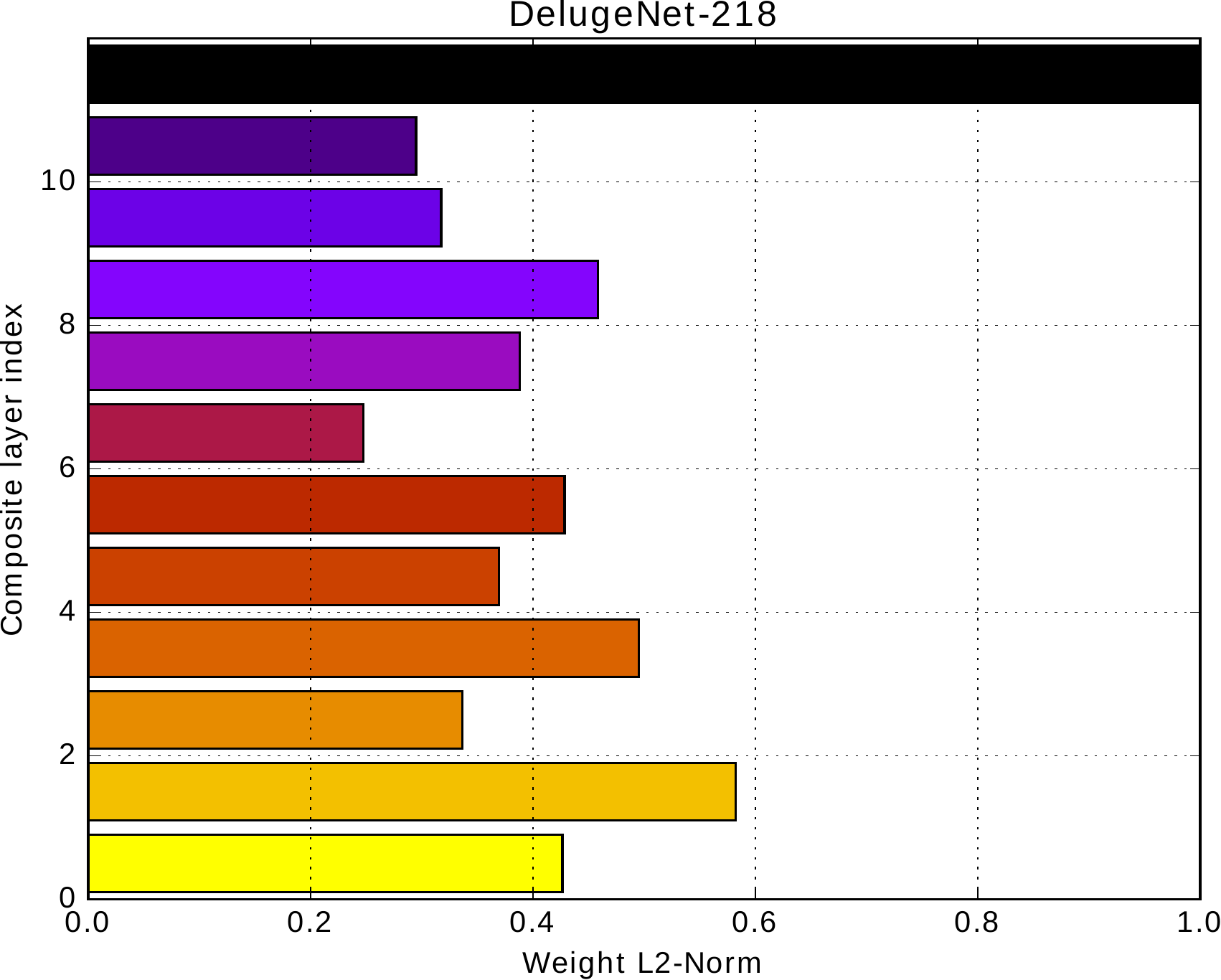} &
		\includegraphics[width=.3\textwidth]{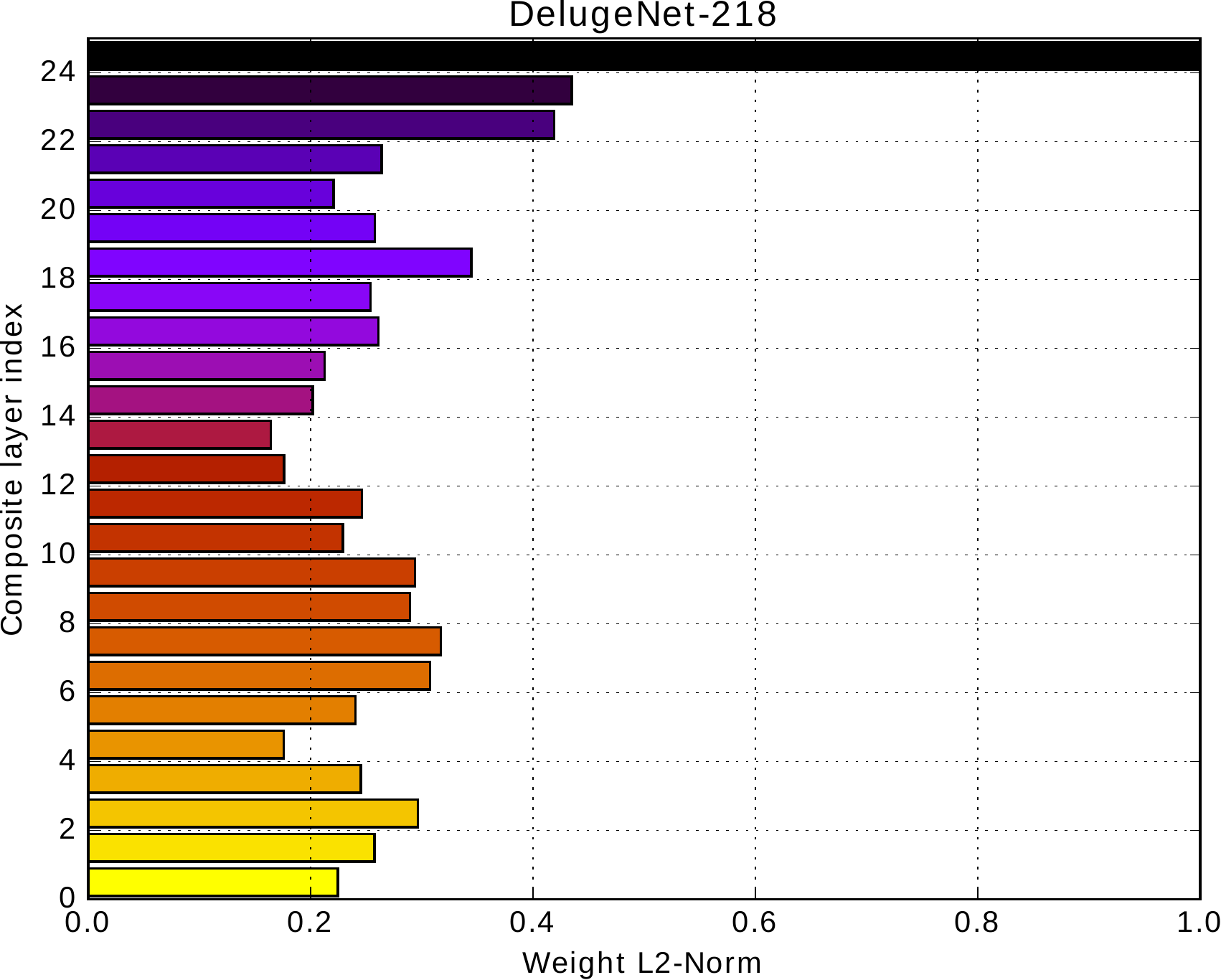} &
		\includegraphics[width=.3\textwidth]{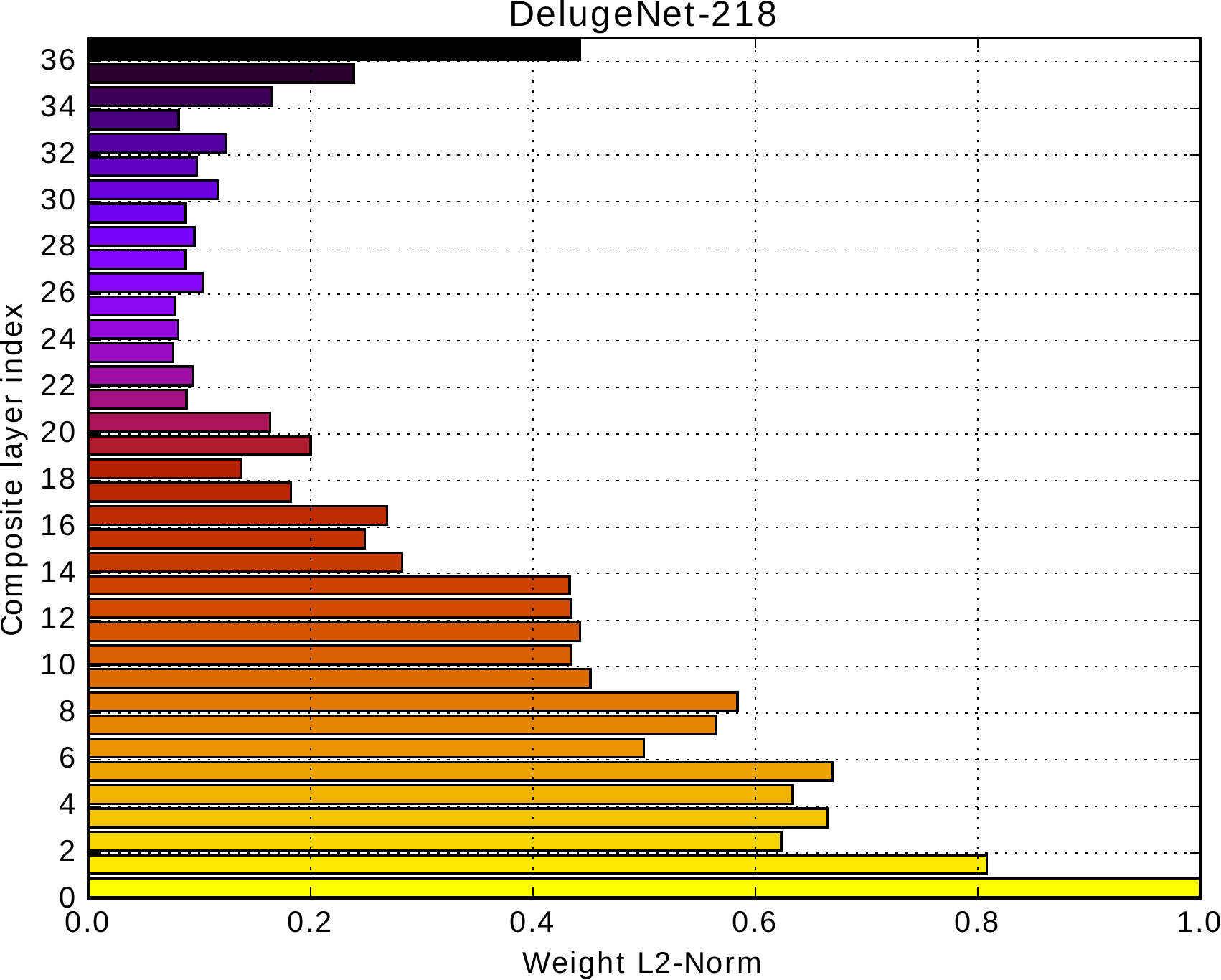} \\
	\end{tabular}
	\caption{Layer-wise L2-norms of cross-layer depthwise convolution weights. Each of the 3 columns corresponds to a different block transition stage in the networks. Vertical axes indicate the indices of the preceding \textit{composite layers}, and horizontal axes indicate normalized L2-norm values. The longer the horizontal bar of a \textit{composite layer}, the larger its contribution.}
	\label{fig:l2norms}
	\vspace*{-0.1cm}
\end{figure*}

\interfootnotelinepenalty=10000
\textbf{Cross-layer connectivity:} For better understanding of cross-layer depthwise convolutional layers, we compute layer-wise L2-norms of the cross-layer depthwise convolutional filter weights of DelugeNet-218, on CIFAR-10 and CIFAR-100. We provide visualizations in Figure \ref{fig:l2norms}. The weight's L2-norms are normalized\footnote{The relative (as opposed to absolute) L2-norm values are sufficient, since BN layers follow cross-layer depthwise convolutional layers.} by dividing them with the maximum layer-wise L2-norms of every block. We consider only cross-layer depthwise convolutional layers in the \textbf{Block Transition 1} (from Block 1 to Block 2), \textbf{Block Transition 2} (from Block 2 to Block 3), and the cross-layer depthwise convolutional layer (from Block 3 to classification) \textbf{before classification layer}. These are the cross-layer depthwise convolutional layers with the highest numbers cross-layer connections in the networks.

Generally, all of the preceding \textit{composite layers} contribute reasonably, with a few dominating. The weights (initialized uniformly) are no longer uniform for all layers in the trained models, being different from the connection rigidity exhibited by ResNets. For first and second block transitions, the last \textit{composite layers} always contribute the most, somehow approximating the behaviors of conventional neural networks where all incoming information comes solely from the layer just before the current layer. On the other hand, for the cross-layer depthwise convolutional layer (before classification layer) connected to the third network block, the early \textit{composite layers} generally contribute the most, and the final \textit{composite layer} contributes moderately. We reckon that the features computed by the earlier \textit{composite layers} are fairly ready for classification, and the subsequent \textit{composite layers} just refine them further. Such phenomenon has also been observed in ResNets \cite{veit2016residual}, where upper layers could be deleted without hurting performance much. In addition, we notice that some \textit{composite layers} in the first block of DelugeNet-218 (CIFAR-100) hardly have any contributions to Block Transition 1. This observation may suggest that \textit{layer sparsity} can be potentially exploited for training DelugeNets.

\subsection{ImageNet}\label{imagenet}
\textbf{Dataset:} ImageNet (1000 classes) dataset \cite{russakovsky2015imagenet} is the most widely used large-scale image classification dataset in recent years. We report the results for validation images. We follow the data augmentation scheme adopted in GoogleNet/Inception \cite{szegedy2015going,szegedy2016rethinking} and ResNet-v2 \cite{he2016identity} with the following augmentation techniques: scale \cite{krizhevsky2012imagenet} \& aspect ratio \cite{szegedy2015going} augmentation, PCA-based lighting augmentation \cite{krizhevsky2012imagenet}, photometric distortions \cite{howard2013some}, and horizontal flipping. The images are normalized by subtracting them from channel-wise means and dividing them by channel-wise standard deviations.

\textbf{Implementation:} We implement and evaluate 3 different DelugeNet models on ImageNet dataset. Similar to ResNet models \cite{he2016identity}, before being passed to the first block, the feature map (after first layer) is downsampled to spatial dimensions of 56$\times$56 via max-pooling. We set the base feature channel dimensions (widths) of all ImageNet-based DelugeNet models to be identical to those of ResNets’ \cite{he2016identity} - \{64,128,256,512\}. Most of the network architectural details follow ResNets' closely, and they are not necessarily optimal for DelugeNets. Moreover, we emphasize on great simplicity when choosing the \textit{composite layer} counts for DelugeNets, setting the number of \textit{composite layers} in each block to be larger than or equal to that of its preceeding block. This is in contrast to the carefully tuned \textit{composite layer} counts \cite{he2016identity} (e.g., \{3,4,23,3\}, \{3,8,36,3\}) in ResNets. The specifications of DelugeNet models are as follows:

\textbf{DelugeNet-92} has \textit{composite layer} counts of \{7,7,8,8\}, for its 4 blocks (in sequential ordering) respectively. \textbf{DelugeNet-104} and \textbf{DelugeNet-122} are two deeper DelugeNet models, with \textit{composite layer} counts of \{7,8,9,10\} and \{7,9,11,13\} respectively.

The ImageNet-based models are initialized similarly to the CIFAR models. Training is carried out with SGD over a total of 100 training epochs, with Nestorov Momentum \cite{sutskever2013importance} and weight decay rate of $1\mathrm{e}{-4}$. We start with learning rate of 0.1, and decay it by factor of 0.1 at the end of every thirty epoch. The training mini-batch size is 256. In view of large model and image sizes, we train the models in multi-GPU mode with 8 GPUs, splitting each mini-batch into 8 portions. These are standard training settings and similar to those \cite{facebook2016resnet} used to train ImageNet-based ResNets.

\textbf{Results:} The top-1 and top-5 classification errors achieved by DelugeNets on ImageNet validation dataset are presented in Table \ref{table:imagenet}, along with the numbers of floating-point operations (GigaFLOPs/GFLOPs) required by the models to process one image. For comparison, we include the results of ResNet-v2 \cite{he2016identity}, Wide-ResNet \cite{zagoruyko2016wide}, and DenseNet \cite{huang2016densely}.

DelugeNet-92 with merely 43.4M parameters outperform ResNet-101 (top1 +0.39\%, top5 +0.18\%) and even ResNet-152 (top1 +0.11\%, top5 +0.13\%). Besides, with 25.5M less parameters and about half (11.8 GFLOPs) of the Wide-ResNet-50's FLOPs (22.8 GFLOPs), DelugeNet-92 performs comparably to Wide-ResNet-50. Both deeper models DelugeNet-104 and DelugeNet-122 further push down the classification errors substantially. Remarkably, DelugeNet-122 attains classification errors comparable to ResNet-200's, despite needing just about half (15.2 GFLOPs) of the computations required by ResNet-200 (30.1 GFLOPs). With the flexible cross-layer connections established by cross-layer depthwise convolutions, DelugeNet-122 is robust against the overfitting issue caused by allocating more parameters to the spatially smaller blocks. Moreover, DelugeNet-122 outperforms DenseNet-161 (best DenseNet model reported for ImageNet dataset) given similar model complexities.

Given similar or considerably lower model budgets (GFLOPs, number of parameters), DelugeNets are able to surpass ResNets, although DelugeNets' \textit{composite layer} counts are configured in a rather simple manner.

\begin{table}[t]
	\centering
	\begin{center}
		\resizebox{\columnwidth}{!}{%
			\begin{tabular}{l|r|r|r|r}
				\textbf{Model} & \textbf{\#Params} & \textbf{GFLOPs} & \textbf{Top-1} & \textbf{Top-5} \\ \hline
				ResNet-101 \cite{he2016identity}    & 44.6M    &   15.7  & 22.44           & 6.21           \\
				ResNet-152 \cite{he2016identity}    & 60.3M    &  23.1   & 22.16           & 6.16           \\
				ResNet-200 \cite{he2016identity}   & 64.8M      & 30.1     & 21.66        & \textbf{5.79}           \\ \hline
				Wide-ResNet-50 \cite{zagoruyko2016wide}   & 68.9M      & 22.8     & 21.9        & 6.03           \\ \hline
				DenseNet-161 \cite{huang2016densely}   & 28.7M      & 15.5     & 22.2        & -           \\ \hline
				\textbf{DelugeNet-92}  & 43.4M    &  11.8   & 22.05  & 6.03       			 \\
				\textbf{DelugeNet-104} & 51.4M    &  13.2   & 21.86  & 5.98           	 \\
				\textbf{DelugeNet-122} & 63.6M    &  15.2   & \textbf{21.53}  & 5.86           	 \\ \hline
			\end{tabular}
		}
	\end{center}
	\caption{ImageNet validation errors (single 224$\times$224 crops).}
	\label{table:imagenet}
	\vspace*{-0.1cm}
\end{table}

\section{Memory Usage}
In ResNets, the residual (addition) operation allows memory buffers to be shared or reused across consecutive composite layers. However, for DelugeNets and DenseNets \cite{huang2016densely}, the output activations and gradients of the last convolutional layer of every composite layer have to be retained persistently during training. For instance, when doing training (\textbf{\^{T}}) and inference (\textbf{\^{I}}) with Wide-DelugeNet-146 on CIFAR-100 (batch size of 32), the occupied GPU memory is roughly \{\^{T}: 2.8G, \^{I}: 0.61G\}, while its ResNet-baseline counterpart only requires \{\^{T}: 1.3G, \^{I}: 0.44G\}. The gap is smaller during inference ($1.5\times$) than in training ($ 2.2 \times$). On the other hand, DenseNet($k=24$), DenseNet-BC($k=24$), and DenseNet-BC($k=40$) require \{\^{T}: 6.3G, \^{I}: 0.87G\}, \{\^{T}: 8.6G, \^{I}:0.75G\}, and \{\^{T}: 8.4G, \^{I}: 1.1G\} respectively. Wide-DelugeNet-146 is more memory-efficient than DenseNet($k=24$), while DenseNet-BCs are very memory-costly.

\section{Conclusion}
We extend depthwise convolutional layers to \textit{cross-layer depthwise convolutional layers}, which facilitate cross-layer connections in our proposed DelugeNets. The cross-layer information inflows in DelugeNets are \textbf{flexible} (cross-layer depthwise convolutional filters are learned) yet \textbf{massive} (output widths of composite layers are regular). Experiments indicate that DelugeNets are quite comparable to state-of-the-art models in terms of accuracies, and yet DelugeNets have lower model complexities. This suggests that DelugeNets may have potentials in energy-efficient deep learning. In future, we would like to investigate regularization techniques (e.g., layer dropout \cite{huang2016deep}) in the context of cross-layer connectivity, as well as applying DelugeNets to other vision applications.

\section{Acknowledgement}
This research was carried out at the Rapid-Rich Object Search (ROSE) Lab at Nanyang Technological University (NTU), Singapore. ROSE Lab is supported by the National Research Foundation, Singapore, under its Interactive Digital Media (IDM) Strategic Research Programme. We gratefully acknowledge the GPU resources and support provided by NVAITC (NVIDIA AI Technology Centre) Singapore.

{\small
\bibliographystyle{ieee}
\bibliography{egbib}
}

\end{document}